\pdfoutput=1
\documentclass[11pt]{article}
\usepackage{tabularx}
\usepackage[]{acl}
\usepackage{array}
\usepackage{times}
\usepackage{latexsym}
\usepackage{multirow}
\usepackage{adjustbox}
\usepackage{tabularray}
\usepackage{float}
\usepackage{graphicx}
\usepackage{epstopdf}
\graphicspath{figures/}
\newcommand\T{\rule{0pt}{2.6ex}}       % Top strut
\newcommand\B{\rule[-1.2ex]{0pt}{0pt}} % Bottom strut
\usepackage{titlesec}
\usepackage[T1]{fontenc}
\usepackage[utf8]{inputenc}
\usepackage{microtype}

\usepackage{inconsolata}

\title{Diverse Perspectives, Divergent Models: Cross-Cultural Evaluation of Depression Detection on Twitter}

\author{Nuredin Ali$^{1}$, Charles Chuankai Zhang$^{1}$, Ned Mayo$^{2}$ , Stevie Chancellor$^{1}$ \\
  \\
  $^{1}$University of Minnesota \\
  \texttt{\{ali00530,zhan6914,steviec\}@umn.edu} \\
  $^{2}$Macalester College \\
  \texttt{emayo@macalester.edu}
}
\begin{document}
\maketitle
\begin{abstract}
Social media data has been used for detecting users with mental disorders, such as depression.\ Despite the global significance of cross-cultural representation and its potential impact on model performance, publicly available datasets often lack crucial metadata related to this aspect.\ In this work, we evaluate the generalization of benchmark datasets to build AI models on cross-cultural Twitter data. We gather a custom geo-located Twitter dataset of depressed users from seven countries as a test dataset\footnote{Details of the cross-cultural evaluation dataset used in this work: {https://grouplens.org/datasets/twitter-depression-dataset-2024/}}. Our results show that depression detection models do not generalize globally. The models perform worse on Global South users compared to Global North.\ Pre-trained language models achieve the best generalization compared to Logistic Regression, though still show significant gaps in performance on depressed and non-Western users.\ We quantify our findings and provide several actionable suggestions to mitigate this issue.
% Additionally, we propose suggestions to improve the generalizability of existing models from social media for future research on mental health detection.   
\end{abstract}

\section{Introduction}
According to the data from World Health Organization, depression is a global issue affecting 240 million people worldwide\footnote{ https://vizhub.healthdata.org/gbd-results/}. In response to these trends, in the last decade, there has been a surge in studying the mental health status of users from social media based on their content and interaction~\cite{ji2018supervised}. Research has focused on various disorders, including depression, anxiety, and eating disorders, and has used many methods~\cite{wongkoblap2017researching}. Specifically - depression is among the most widely studied disorders (and the most commonly diagnosed), and Twitter is a common source of data in these studies~\cite{chancellor2020methods}. 
\begin{figure}[ht]
  \centering
  \includegraphics[width=0.5\textwidth]{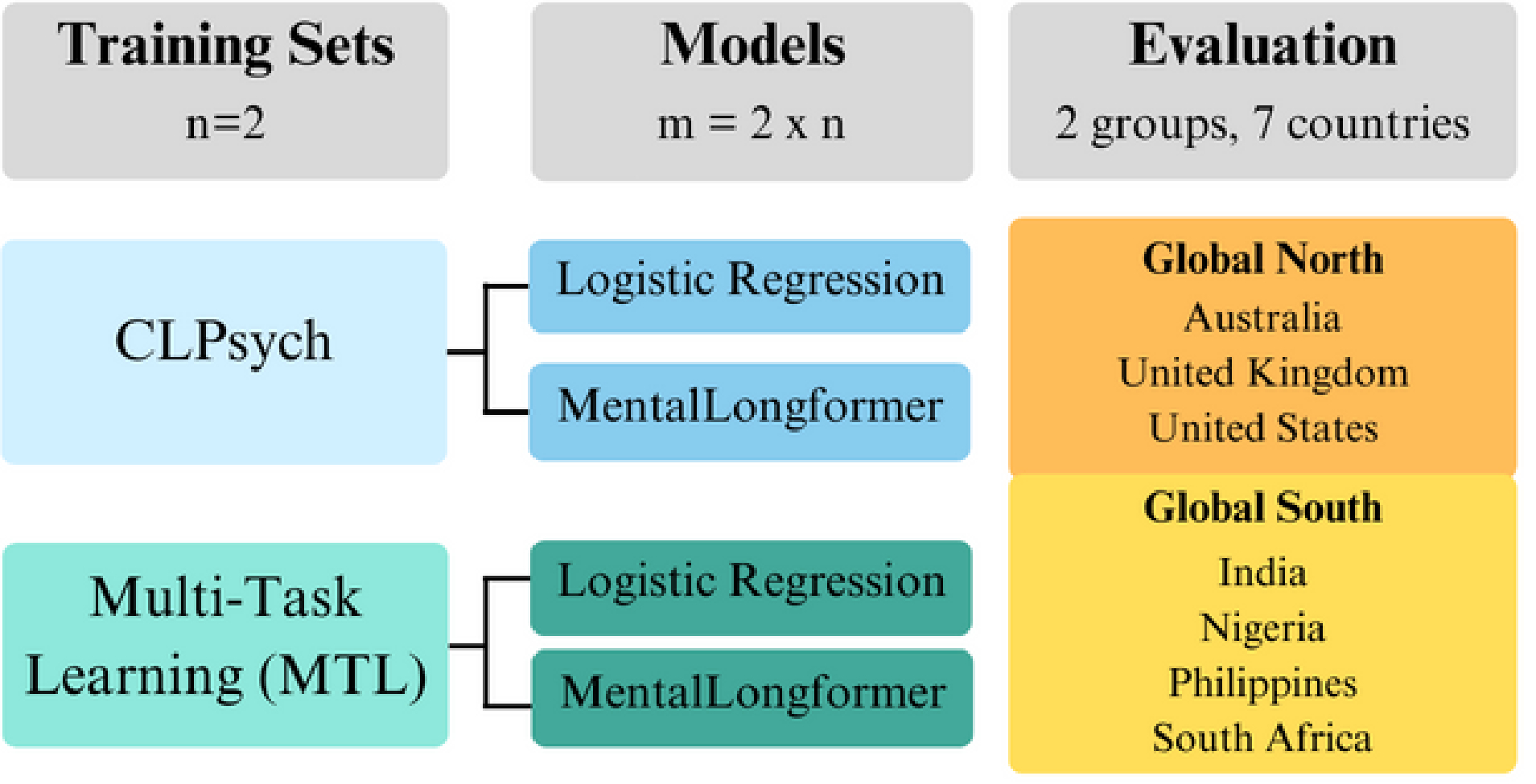}
 \caption{Flow chart of the overall design of the work. This shows the training and evaluation process. n=datasets, m=models.}\label{fig 1:}
 \vspace{0pt}
\end{figure}
This work tries to predict if someone may have depression based on data from social media~\cite{chancellor2020methods, harrigian2021state}. 
%Institute of Health Metrics and Evaluation. Global Health Data Exchange (GHDx).  https://vizhub.healthdata.org/gbd-results/ (Accessed 4 March 2023).}

% cross-cultural studies: 

Given this area's popularity and potential reach to clinical settings, NLP has also called for careful evaluations of bias, performance gaps, and generalizability of claims from small datasets~\cite{aguirre2021gender, harrigian2020models, hovy2016social}. One source of underexplored bias in these datasets and models is the impact of a person's geographic location (and consequently, their culture) on their communication style. The importance of cultural consideration in social media studies about mental health is critical~\cite{lee2014social}. In prior work,~\citeauthor{de2017gender} showed cross-cultural differences in mental health communication styles in cultures such as the US, India, and the Philippines. Cross-cultural users also have different identity dimensions, language use, and support behavior~\cite{pendse2019cross, mittal2023language}; sentiment detection can vary across cultures~\cite{pruksachatkun2019moments}.\ However, interaction in international forums does not affect their clinical mental health language use~\cite{pruksachatkun2019moments}.\ Recent literature reviews~\cite{chancellor2020methods} and persuasive calls~\cite{garg2023mental} point to the need to study the generalizability of models to distinctive user populations for mental health research. Similar audits have been instrumental in identifying gaps in performance across clinical/non-clinical populations~\cite{10.1145/3290605.3300364} and in gender and racial groups~\cite{harrigian2020models, aguirre2021gender, aguirre2021qualitative}.
%\cite{de2017gender} analyzing depression across cultures, found a difference in expression.  These studies conducted within two or three cultures focusing on the linguistic analysis showed users from different cultures express their mental health status differently. Building on this line of research, this work analyzes the generalization of depression detection models on cross-cultural data trained on existing benchmark datasets.

Building on this prior research, in this paper, we analyze the generalization of depression detection models on cross-cultural data trained on existing benchmark datasets. Inspired by~\cite{harrigian2020models,aguirre2021gender}, we ask: do models built on popular social media benchmark datasets to predict depression generalize to people who live in different countries, yet speak English? If the prior work is correct about geographic and cultural biases impacting predictions, what countries may be most affected? How stark are the performance differences between countries? 

We audit depression detection models generalization on cross-cultural data gathered from Twitter (now X). We collected data from seven countries using a strict location verification technique and the prevalence of English content in users' feeds, using keyword matching and manual annotation to identify genuine depression disclosures. We trained two models, Logistic Regression and MentalLongformer, on benchmark depression datasets (CLPsych and Multi-Task Learning). We assessed their generalization by both country and socio-economic development classification (Global North vs. Global South).

We show that models trained on broad Twitter benchmarks do not generalize well to the cross-cultural data. Models generalize much better to evaluation data from users in the Global North (US, UK, Australia) than to users in the Global South that use English as a national language (India, Nigeria, Philippines, South Africa). Distinct gaps emerge between countries, with models generalizing very poorly to posters from Nigeria and India. 
% Relatively, MentalLongerformer performed the best trained on the Multi-Task Learning data. 
%Both models trained on either data generalize poorly to Nigeria and India. There's a higher disparity in performance between the Global North and Global South countries. 
Our findings demonstrate that existing benchmark datasets are not representative of training generalized models that could detect depressed users from various cultures. We provide suggestions for building better datasets and models.

\section{Datasets} \label{datasets_section}

%Assume the reader is dumb - remind them what the problem/goal is of the paper here before you explain how we got the dataset. Two sentences will be fine.
%Current widely-used benchmark datasets do not include user geo-location meta-data, preventing the understanding of how various methods generalize across cultural contexts. 
We carefully selected two popular benchmark datasets for constructing depression models on Twitter data and then created a geolocated dataset of depression posts. Table~\ref{tab:datasets} summarizes our datasets.

\begin{table}
\small
\centering
\resizebox{0.5\textwidth}{!} {
\begin{tabular}{llcr} 
%\hline
\textbf{Dataset} & \textbf{Classes} & \textbf{Train} & \textbf{Val}\\ 
\hline

CLPsych  & Depression & 327 & 150 \\
 \cite{coppersmith2015clpsych} & Control & 570 & 301 \\
\hline
Multi-Task Learning  & Depression & 1520 & 320 \\
\cite{shen2017depression} & Control & 1520 & 320 \\
\hline
Ours (Evaluation) & Depression &   - & 267 \\
 & Control &   - & 264 \\
\end{tabular}
}
\caption{Datasets used in our experiments.}
\label{tab:datasets}
\end{table}

\medskip
\noindent\textbf{CLPsych}: This dataset comes from the CLPsych 2015 Shared Task~\cite{coppersmith2015clpsych}. The shared task contains two mental disorder identifications, identified with keywords and manual annotation: depression and PTSD, of which we use the depression treatment data and control. The data comprises the users' most recent posts around the date of depression disclosure, up to a maximum of 3000 posts per user. 
% The shared task contains Twitter user timelines about two conditions: depression and PTSD. Each user's dataset includes up to their most recent 3000 public tweets. Human annotators evaluated each segment to include only genuine disclosures of a mental health condition. For this work, we consider the depression condition only and their control counterparts.     

\medskip
\noindent\textbf{Multi Task Learning (MTL)}: This dataset is from~\cite{shen2017depression}, which contains Twitter user profile information and their posts within one month. This dataset also identified people who may be depressed in Twitter. Both CLpsych and MTL are gathered based on a strict set of keywords/keyphrases such as “(I’m/I was/ I am/ I’ve been) diagnosed depression,” etc., to identify the candidate depressed users. We leverage the text data only (as this dataset does contain images).

% After preprocessing, the final training set contained 1520 depression users, and the test set contained 320 users. As this dataset contains images associated with some posts, we ignore the images to look only at the text. Both the CLPsych and the MTL datasets are gathered based on a strict set of keywords/keyphrases such as “(I’m/I was/ I am/ I’ve been) diagnosed depression,” etc., to identify the candidate depressed users.

\medskip
\noindent\textbf{Our global dataset}:
At the time of writing, there are no public benchmark datasets of global expressions of depression in Twitter data. Therefore, we collect a corpus from public posts from Twitter using the Twitter Research API (now defunct)\footnote{https://developer.twitter.com/en/use-cases/do-research/academic-research}. We used the search terms/phrases from~\citeauthor{de2017gender}'s cross-cultural depression study on Twitter to identify people discussing depression or suicidality (a common co-morbid symptom of depression). These include phrases such as ``I am/I'm depressed'' and ``I want to hurt myself'', and were verified by psychologists by the collaborators of~\citeauthor{de2017gender} We searched the sample of the Twitter data made available between January 2015 and December 2022\footnote{Note that the Twitter API gave a sample of data, but not all of it.}.
%Due to various competitive advantages, social media platforms, including Twitter (X), do not extend access to their full data. \cite{acker2020social} }.  

% The 3200 posts before the disclosing post have been collected to get the user's history. The limit is applied due to Twitter's API restriction.  
 
Given our focus on cross-cultural content, we leveraged geotagged tweets. We specifically gathered users from seven countries: Australia, South Africa, Nigeria, the Philippines, India, the United Kingdom, and the United States. We selected these countries for geographic diversity, their large volume of geotagged disclosures, and the fact that English is a first language or is a business/government-listed language in those countries. To verify that a user was in the country, we looked at their 3200 geotagged posts before disclosure and took the country where the user posted the most. %We combine this history of posts in our data. 

We manually verified each user's veracity of depression disclosure with human raters, similar to the process in \cite{coppersmith2015clpsych}. We developed and applied a codebook to identify users who had genuine disclosures of depression (see Appendix \ref{human_verification} for details). We then gathered control users with similar demographics whose posts did not include the search terms but with the same geotagged Tweet rules as discussed. Therefore, we made a matched, "control" sample of users from the same country who disclosed having depression and those who did not. 

% We gathered each user's tweet history before the disclosure using either the matched disclosure or control post as a starting point. We gathered the 3200 posts preceding the disclosure post for each user, aligning with the posts used for geo-location verification. We then manually verified each user's veracity of depression disclosure with human raters, similar to the process in\cite{coppersmith2015clpsych} (see Appendix \ref{human_verification} and \ref{gen_vs_non_examples} for details). We gathered control users with similar demographics whose posts did not include the search terms but with the same geotagged Tweet rules as discussed.

At the outset, our dataset comprised 16,112 potentially depressed users from the seven countries. Of these, 1,556 were manually reviewed, leading to the identification of 267 authentic disclosures. The annotation encompassed all original sample users from the Global South countries but did not cover all users from the Global North due to the substantial volume of data. Cohen's Kappa~\cite{mchugh2012interrater} between the raters resulted in 0.65, showing a substantial agreement. The disagreements were resolved through two rounds of discussion. % Table \ref{tab:genuine_examples}% 
Appendix \ref{gen_vs_non_gen} shows genuine and non-genuine posts obtained through human annotation. These examples are paraphrased and lightly edited to protect the identity of the posters~\cite{ayers2018don}.

Our geo-located dataset encompasses a total of 531 users, with 267 users identified as having depression through manual verification. We define two groups of countries based on the United Nations categorization of countries\footnote{https://unctad.org/system/files/official-document/tdstat47\_en.pdf} - the Global North and the Global South. 140 users were in the Global North (64 United States, 45 United Kingdom, 31 Australia), while 127 users hail from the Global South (58 Philippines, 35 South Africa, 19 India, 15 Nigeria). 

% After applying the preprocessing, the final evaluation set includes 112 depressed users, 16 from each country, to take the balanced version. 

\subsection{Preprocessing}
We applied the same preprocessing pipeline across the datasets (including the benchmarks) for consistency, following recommendations from \cite{harrigian2020models}. 
Specific retweet tokens, username mentions, URLs, and numeric values were removed. English contractions were expanded. We removed the disclosure words from the training and evaluation sets. Users with fewer than 20 posts were excluded, and only those with a minimum of 20 English posts were considered for inclusion.

\section{Baseline Models}

%We first reviewed the depression detection literature that introduced different methods using these text corpora for each dataset in detecting depressed users. Through this review, we identified the baseline models. 

For mental health prediction tasks, Logistic Regression is a popular and performant statistical baseline due to its quick training time, success in prediction, and highly interpretable feature relevance~\cite{benton-etal-2017-multitask, jiang2018detecting, harrigian2020models}. In this experiment, we extracted the features using the term frequency-inverse document frequency (TF-IDF). Using scikit-learn, we applied grid search on 5-fold cross-validation. The best hyperparameters for the logistic regression are \textit{'penalty': 'l2', 'solver': 'lbfgs',  'max\_iter': 10000}, and 7000 TF-IDF features.

\begin{table}[ht]
\vspace{-5pt}
\small
\centering
\setlength\extrarowheight{2pt}
% \large
\resizebox{0.5\textwidth}{!} {
\begin{tabular}{lcccr}
\multirow{2}{*}{Model} & \multicolumn{2}{c}{CLPsych} & \multicolumn{2}{c}{MTL} \\
\cline{2-3}\cline{4-5}
& Recall & F1 & Recall & F1 \\ 
\hline
Logistic Regression & 0.83 & 0.80 & 0.89 & \textbf{0.89} \\ 
MentalLongformer & 0.76 & 0.73 & 0.93 & \textbf{0.92} \\
\end{tabular}
}
\caption{The F1 score of the baseline model on both datasets.}
\vspace{-5pt}
\label{tab:baseline_results}
\end{table}

Pretrained language models such as BERT (Bidirectional Encoder Representations from Transformers)~\cite{devlin2018bert} have significantly improved text classification on many general~\cite{murarka2020detection} and domain-specific tasks~\cite{ji2021mentalbert,ji2023domain}. We finetune the MentalLongformer language model for our baseline, which outperforms the other pre-trained models on this specific task and has an extended sequence modeling capacity~\cite{ji2023domain}. We use it to investigate its generalization capabilities to our task.

\begin{table*}[]
\centering
    \small 
    %\hline 
    \resizebox{1.0\textwidth}{!} {
\begin{tabular}{lcc|cc|cc|cc|cc|cc|cr}
\T
\multirow{2}{*}{Training Data}                                                                                                                                                                                    
                               & \multicolumn{2}{c|}{Australia} & \multicolumn{2}{c|}{Nigeria} & \multicolumn{2}{c|}{South Africa} & \multicolumn{2}{c|}{Philippines} & \multicolumn{2}{c|}{India} & \multicolumn{2}{c|}{UK} & \multicolumn{2}{c}{US} \\ \T \B
                               % \hline \\
                               & Recall         & F1           & Recall        & F1          & Recall           & F1            & Recall          & F1            & Recall       & F1         & Recall            & F1             & Recall           & F1             \\ 
                               \hline
                               \T
CLPsych                        & 0.61           & 0.63         & 0.13          & 0.23        & 0.45             & 0.53          & 0.39            & 0.46          & 0.10         & 0.19      & 0.53              & 0.61           & 0.53             & \textbf{0.66}           \\ \B \B \T
Multi Task Learning            & 0.53           & 0.60        & 0.13          & 0.23        & 0.28             & 0.36          & 0.08            & 0.15          & 0.26         & 0.35       & 0.84              & \textbf{0.69}           & 0.75             & 0.61          
\end{tabular}
}
\caption{F1 scores of \textit{Logistic Regression} trained on CLPsych and Multi-Task Learning datasets.}
    \label{tab:lr_results}
\end{table*}

For this experiment, the pre-trained head of the MentalLongformer is replaced with a randomly initialized classification head. We set the learning rate to 5e-5. Adam is used as an optimizer \cite{kingma2014adam}. We trained for ‘num\_train\_epochs=50’ and applied an ‘early\_stopping\_patience=10’. The remaining parameters were set to the default hyperparameters of MentalLongformer on Huggingface. Table~\ref{tab:baseline_results} presents the results of the baseline models on the test of both datasets. To evaluate the model's performance, we report F1 and recall, selected for their effectiveness in handling unbalanced datasets. %The Global North and Global South sets comprise similar user groups despite the varying proportions across countries.

\begin{table*}[ht]
\centering
    \small 
    %hline 
    \resizebox{1.0\textwidth}{!} {
\begin{tabular}{lcc|cc|cc|cc|cc|cc|cr}
\T
\multirow{2}{*}{Training Data}                                                                                            
                               & \multicolumn{2}{c|}{Australia} & \multicolumn{2}{c|}{Nigeria} & \multicolumn{2}{c|}{South Africa} & \multicolumn{2}{c|}{Philippines} & \multicolumn{2}{c|}{India} & \multicolumn{2}{c|}{UK} & \multicolumn{2}{c}{US} \\ \T \B
                               % \hline \\
                               & Recall         & F1           & Recall        & F1          & Recall           & F1            & Recall          & F1            & Recall       & F1         & Recall            & F1             & Recall           & F1             \\ 
                               \hline
                               \T
CLPsych                        & 0.64           & \textbf{0.72}         & 0.06          & 0.12        & 0.2             & 0.32          & 0.18            & 0.3          & 0.15         & 0.27       & 0.37              & 0.5           & 0.42             & 0.56           \\ \B \B \T

Multi Task Learning            & 0.93           & 0.69         & 0.33          & 0.45        & 0.71             & 0.67          & 0.31            & 0.43          & 0.68         & 0.7       & 0.95              & \textbf{0.72}           & 0.84            & 0.64          
\end{tabular}
}
    \caption{F1 scores of \textit{MentalLongformer} trained on CLPsych and Multi-Task Learning datasets.}
    \label{tab:mentalformer_results}
    
\end{table*}

%%% RESULTS GLOBAL NORTH VS SOUTH

\begin{table}[ht]
\centering
    \small
    \resizebox{0.5\textwidth}{!} {
\begin{tabular}{lc|cc|cr}

\T 
\multirow{2}{*}{Model}               & \multirow{2}{*}{Training set} & \multicolumn{2}{c|}{Global North} & \multicolumn{2}{c}{Global South} \\ \B
                                     &                               & Recall           & F1            & Recall           & F1            \\ \hline \T \T 
\multirow{2}{*}{Logistic Regression} & CLPsych                       & 0.47             & \textbf{0.58}          & 0.17             & 0.28          \\ 
                                     & Multi-Task Learning           & 0.76             & \textbf{0.63}          & 0.17             & 0.26          \\ \T
\multirow{2}{*}{MentalLongformer}    & CLPsych                       & 0.45             & \textbf{0.58}          & 0.17             & 0.28          \\ 
                                     & Multi-Task Learning           & 0.9              & \textbf{0.68}          & 0.48             & 0.56         
\end{tabular}
}
\caption{F1 scores of both models trained on CLPsych and Multi-Task Learning datasets evaluated on Global North and Global South eval sets.}
    \label{tab:global_n_s}
    \vspace{-7pt}
\end{table}

\section{Results}

In Table~\ref{tab:baseline_results}, we present the results of our ML models on two benchmark datasets (CLPsych and MTL). Our baseline models closely replicate prior research of benchmark datasets (logistic regression~\cite{aguirre-etal-2022-basco}, and MentalLongformer~\cite{ji2023domain}). We evaluate model performance on our custom dataset, split into two groups - Global North vs. Global South and then country-level. %We report these baseline models' Recall and F1 scores in the test set in Figure \ref{fig 1:}. Note: in this analysis, the depression users are referred to as positive.

% [a single summary sentence of the BLR + cites to what the baseline is]. [a single summary sentence of the transformer + cites to the baseline]. 

\subsection{Global North vs. Global South}
% [SC: describes results of our study when looking into the Global North vs. Global South comparisons on our dataset]

Our baseline models trained on benchmark datasets perform much worse on data from the Global South than the Global North. Table~\ref{tab:global_n_s} shows the results of the two groups and our model's performance. There is an expected drop in performance between the baseline model and the Global North and the Global South evaluation datasets (due to them being out-of-domain). However, all four models have a superior F1 and recall in identifying the Global North evaluation users. This finding aligns with prior research that there is a gap in performance between these categories~\cite{pruksachatkun2019moments}, though it confirms it at a larger cultural scale.

Table~\ref{tab:mentalformer_results} shows that the MentalLongformer model trained on the MTL data has much better recall (or sensitivity) for detecting the presence of depression in the Global North countries compared to the Global South. This is particularly useful in these settings where identifying depression users is essential or where models are used for downstream interventions. 
% Even though the MentalLongformer model is better than the Logistic Regression, the drop in recall and F1 are concerning. 
% \vspace{-0.15\baselineskip}
% \titlespacing{\section}{0pt}{\parskip}{-\parskip}
\subsection{Country Level Analysis}
% \vspace{-0.3\baselineskip}
To investigate the performance gap in Global North vs. Global South, we analyze country-level outcomes, presented in Tables~\ref{tab:lr_results} for the Logistic Regression and~\ref{tab:mentalformer_results} for the MentalLongformer model. We separate each country into groups for this analysis, noting that the size of each country's dataset is imbalanced (see Datasets \ref{datasets_section}). 

There is a significant difference (p-value: 0.001) in accurately identifying depressed users among various countries. Further analysis within two groups, (Australia, US, UK) and (India, Nigeria, South Africa, and the Philippines), revealed no statistical differences (p-values: 0.47 and 0.39, respectively). Notably, all models struggled to correctly identify users from Nigeria and India. This disparity indicates the need for more generalizable training benchmarks and models.

%Code-mixed posts from users in Nigeria, the Philippines, and India hamper the detection of depressed users from these countries. Refer to \ref{error_analysis} for examples.  

% [SC: Then, country specific analysis - this is a good section to describe where it was good/bad and why - you may want to look at confusion matrices, examples of misclassification, or other "evidence" that can explain what may be going on that causes the misclassification]

\subsection{Qualitative Error Analysis} \label{error_analysis}

To understand the disparities in detection, we explore the model with the highest variance in F1 score between the Global North and Global South (the Logistic Regression model trained on the MTL dataset, with a 0.37 F1 score gap between these two regions). We conducted a qualitative error analysis focusing on users from Nigeria and the Philippines. We initially look at the word distributions between these two regions to discern potential similarities/differences in the most frequent words. We present a few qualitative observations of trends.

First, users in the Global South, particularly those from Nigeria and India, express common words such as 'god,' 'life,' 'love,' and 'people.' Within Global North countries, words like 'work,' 'day,' 'time,' and 'people' rank prominently among common words. There are shared linguistic features between both regional sets of countries, such as 'love,' 'life,' 'like,' and 'one.’ Still, we note that many of the users from Nigeria and India have discrepancies in how they communicate in general (not just about mental health). This aligns with prior research that highlights variations in linguistic patterns~\cite{pendse2019cross,de2017gender}. Such differences in language usage might account for the subpar performance observed across these countries. 

Second, we also note that some users in non-Western countries rarely engage in code-mixing, where they use two or more languages in speech at a given time. Take this example user, who contains both English and other languages that the model misclassifies.- (e.g. \textit{‘here it goes no one wants me i am worthless even though i am alive feeling dead inside gusto ko magbakasyon ng mahabang mahaba...’}. Similar trends happen in Nigerian users, e.g. \textit{‘wani abu ma sai dan shaye shaye sai ma lokacin iftar zata ga abubuwa amin lemme just pretend i did not see that’}. However, there are no such examples of code-mixing in the Global North countries where English is the primary official language. Recall that we picked countries where English is an official language or would be used in business settings and identified Twitter users who primarily Tweeted in English. However, identifying code-mixed tweets is challenging, and Twitter’s language detector has limitations.

\section{Recommendation and Conclusion}

In this work, we quantified the generalization capability of depression detection models in cross-cultural data. We specifically quantified that models have higher discrepancies in identifying users from different cultures. We provide the following suggestions for improving the identified gaps.

\textit{Construct datasets with more geographical examples}. Similar to~\cite{harrigian2020models, aguirre2021gender}, we hypothesize that mental health detection from social media suffers from small datasets. Existing benchmark datasets lack the location meta-data of users~\cite{garg2023mental} and lack different demographic representations~\cite{aguirre2021gender}, meaning that fairness audits are challenging to execute post-hoc. 

We propose a few solutions to this problem. First, 
researchers could adapt techniques to infer geo-location if larger datasets were available~\cite{mittal2023language,-2022-international} to conduct audits. Larger datasets could be composited from comparable sources, pointing to evidence from~\cite{harrigian2020models} that more data helps alleviate racial disparities in predictions. Balancing the datasets effectively makes the algorithms fair in different groups~\cite{pessach2023algorithmic}. Ultimately, the field needs to find paths forward to identify and supplement datasets for this task. As an initial stride in this direction, we provide the details for our dataset and how other users may replicate our findings\footnote{The details of the cross-cultural evaluation dataset used in this work is provided {https://grouplens.org/datasets/twitter-depression-dataset-2024/}}
% our dataset will be under IRB review to be made available upon request to other researchers.

%(i.e IBM's AI Fairness 360 balanced facial recognition data highly reduced the disparities in identifying users of different races. \cite{bellamy2018ai}). Algorithmic auditing calls for a rigorous reporting of performance metrics.
    
\textit{Investigating the cross-cultural detection capabilities of proposed models}. Current work has a considerable gap in ethical consideration and transparent reporting~\cite{ajmani2023systematic}. Fine-grained subgroup analysis reporting leads towards building more inclusive and transparent models~\cite{buolamwini2018gender}. We call for critical consideration when reporting these metrics when introducing algorithms.

\section{Ethical Considerations}
Predicting mental health via social media data is ripe with ethical challenges~\cite{benton2017ethical, chancellor2020methods}. Yet, this area also holds promise in identifying early indications of mental disorders, potentially averting risky behaviors, and getting people access to treatment. This requires careful consideration and application in ways that benefit society while mitigating risks.

Our study follows standard procedures for deanonymizing participants in our data~\cite{chancellor2020methods, benton2017ethical}. The IRB at the University of Minnesota (study ID: STUDY00018665) ruled that our work was not human subjects research because our data was publicly available and we did not interact with users. The CLPsych data is accessed through IRB approval, and the Multi-Task Learning data is publicly available. Before computational modeling, we still took procedures to protect participants' identities, such as removing URLs, usernames, and personal identifiers from data. We do not report any data about individuals in certain countries nor provide examples of data to protect people in these situations. It is imperative to underscore that these datasets should exclusively be used for research purposes.

One risk we highlight is cross-cultural factors such as differences in stigma and the consequences of disclosure. Countries have major differences in the stigma and social consequences that the prediction of mental illness may have in those spaces. Individuals can be shamed for disclosing mental illness, prevented from opportunities (employer use in screening processes), or denied dignity. In some cultures, mental illness can be trivialized or ignored. These same factors may lead to different strategies for disclosure in public forums like Twitter. Nonetheless, this data should not be used to draw conclusions about which countries might have higher depression rates or who is ``better'' at caring for people with mental illness. Nor should this data be used to profile people based on inferences from social media data. %We do not have diagnoses for these individuals, an aim of future work.

%The use of social media data to identify mental health issues like depression brings forth substantial data privacy considerations. 
\section{Limitations}

The dataset used for evaluating the six countries might not be representative for three reasons. 

1. Individuals from different countries might convey their mental health status in unique ways, involving using different sets of key phrases compared to those in our study~\cite{pendse2019cross}. To comprehensively understand these potential variations, additional research is required to pinpoint and incorporate these specific keywords and research culture-specific means of disclosure. Moreover, there is also the critical challenge of self-disclosure bias that affects the underlying user sample and modeling output of depressed users \cite{chancellor2023contextual}. 

2. During the qualitative error analysis, we found that users from countries like Nigeria, India, and the Philippines use code-mixing in their posts. Although we filtered for English-only content using Twitter language detection, it missed some posts, resulting in code-mixed content for some users. This could potentially be the source of some of the disparities identified. Therefore, future research could investigate methods to effectively handle code-mixing, enhancing technical capabilities in NLP and cross-cultural mental health detection. 

3. The geo-tagged tweets play a vital role in our research. This constitutes approximately 1\% of Twitter's daily content on Twitter (X) \cite{lamsal2022did}. However, our reliance on this specific subset of data also limits the volume of data in our study.

4. We focused solely on two models, two widely used benchmark datasets, and Twitter (X) as our platform. While this provides valuable insights into disparities, conducting further studies on additional models and platforms could offer a more comprehensive understanding.

\bibliography{anthology,custom}

% \pagebreak

\appendix

\section{Appendix}
\label{sec:appendix}

\subsection{Human Verification of Authentic Mental Health Disclosures} \label{human_verification}

The keyphrases used to search the candidate depression disclosure include words such as \textit{'i [*] diagnosed [*] depression', 'i attempted suicide', 'i am depressed', ' i [have/had] depression', 'i want to die', etc}. 
However, the candidate depression disclosure data is prone to noise. Often, users use these candidate keyphrases in their posts while they are not depressed. For instance, \textit{"I haven’t been to the gym in about a week and a half and I’m depressed."} is not a genuine disclosure according to the annotation rules but would match our keywords. 

We constructed a codebook to manually verify genuine disclosures, building on prior work~\cite{coppersmith2015clpsych}. To classify a post as genuinely about depression, the post must demonstrated that the user states they are sincere about being depressed; a dark joke or sarcasm directly disclosing that they are depressed, suicidal, or thinking about self injury; or the links associated with a post (i.e. images, texts, etc) are related to genuine depression expressions.

Two annotators were involved during the annotation process of the dataset (the first two authors). This includes two PhD students with non-Western backgrounds, and they were supervised by the final author with a Western background and experience in the research area. The two annotators took three rounds of annotation to discuss disagreements on identifying genuine disclosures and refine the process. These discussions were critical to reducing random disagreements~\cite{kapania2023hunt}. The final author consulted on the codebook creation and served as a third deliberation point when needed.

A post is a non-genuine disclosure if the post talks about feelings about a transient situation that uses “depressed” as a stand-in for being sad and the state of mental disorder is unclear, e.g. ‘Manchester United lost the game, I’m depressed.’ or being depressed because you have to go to work when you don’t want to. The majority of posts with language about “being depressed” were ambiguous in these less serious uses of the term depression.

To apply the codebook, we followed the following approach. First, we consulted the post directly to see if it aligned with the codebook. If the post does not provide a full context or was borderline, we looked at the history of the users' posts before the disclosure. If the prior posts do not indicate that the user is depressed, we consider the disclosure as inauthentic. 

\subsection{Distribution of Tokens}

\begin{figure}[ht]
  \centering
  \includegraphics[width=0.5\textwidth]{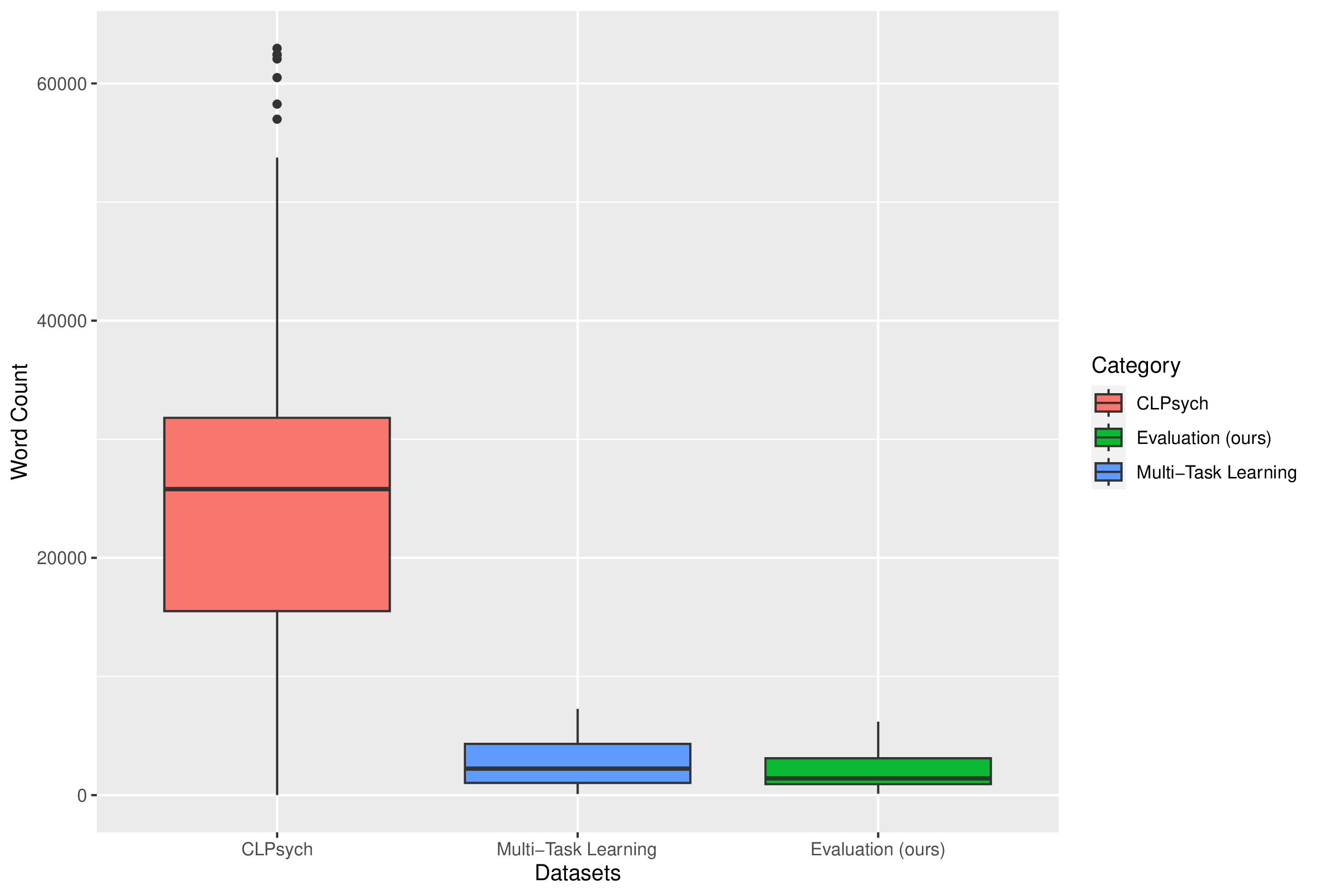}
 \caption{The box plot illustrates the distribution of tokens across the datasets.}
 \label{fig distribution_of_tokens}
\end{figure}

The token distribution differs among the three datasets, with CLPsych containing more tokens than Multi-Task Learning and our dataset, which share a similar proportion see Figure \ref{fig distribution_of_tokens}. However, this variation doesn't significantly impact the models. The MentalLongformer is specifically designed to handle 4096 tokens \cite{ji2023domain}. For logistic regression, we opt for a reduced number of features.

\subsection{Example of Genuine and Non-Genuine Disclosures} \label{gen_vs_non_gen}

\begin{itemize}
    \item \textbf{Genuine Disclosure: } \textit{"His song means more to me now because like I told you I have depression and anxiety. It got so much worse in last few months. Listening that saved me from having a severe mental breakdown and wanting to jump out of my window or do even worse"}, \textit{"I can't pretend to be happy anymore. I cut because I am depressed. I have tried killing myself because I get bullied. I'm not happy."}, \textit{"It's my favorite holiday, and I'm depressed I'm fighting it, but that's exhausting, and so is everything else"}
    \item \textbf{Non-Genuine Disclosure: } \textit{"Haven’t driven my toyota in so longggg. I’m depressed now haha"}, \textit{"These next few months will be dedicated to finally dropping some fucking merch. I've been killing myself over it."}, \textit{"I'm killing myself I'm killing myself I'm killing myself...LoL LoL :-D"}
\end{itemize}

\end{document}